\documentclass[letterpaper, 10 pt, conference]{ieeeconf}  
\IEEEoverridecommandlockouts                              
\usepackage{graphicx}
\usepackage{amsmath} 
\usepackage{amssymb}  
\usepackage{multirow}
\usepackage{algorithm}
\usepackage{algpseudocode}
\usepackage{diagbox}
\usepackage{caption}
\usepackage{paralist}

\long\def\hide#1{}

\title{\LARGE \bf Learning Motion Predictors for Smart Wheelchair using Autoregressive Sparse Gaussian Process
}

\author{
Zicong Fan$^1$, Lili Meng$^1$, Tian Qi Chen$^2$, Jingchun Li$^1$ and Ian M. Mitchell$^1$
\\The University of British Columbia, Vancouver, Canada
\thanks{This work was supported by AGE-WELL NCE Inc. (a member of the Canadian Networks of Centres of Excellence program) and the National Science and Engineering Research Council of Canada (NSERC) Discovery Grant \#298211.}
\thanks{$^{1}$Zicong Fan, Lili Meng, Jingchun Li and Ian M. Mitchell are with the Department of Computer Science at the University of British Columbia in Vancouver, Canada. 
        {\tt\small zfan@alumni.ubc.ca, menglili@cs.ubc.ca, mitchell@cs.ubc.ca }}%
\thanks{$^{2}$Tian Qi Chen is with the Department of Computer Science at the University of Toronto in Toronto, Canada. 
        {\tt\small rtqichen@cs.toronto.edu
}}%
}

\begin{document}
  \maketitle
  \thispagestyle{empty}
  \pagestyle{empty}
  

\begin{abstract}\label{sec:abstract}
Constructing a smart wheelchair on a commercially available powered wheelchair (PWC) platform avoids a host of seating, mechanical design and reliability issues but requires methods of predicting and controlling the motion of a device never intended for robotics.  Analog joystick inputs are subject to black-box transformations which may produce intuitive and adaptable motion control for human operators, but complicate robotic control approaches; furthermore, installation of standard axle mounted odometers on a commercial PWC is difficult.  In this work, we present an integrated hardware and software system for predicting the motion of a commercial PWC platform that does not require any physical or electronic modification of the chair beyond plugging into an industry standard auxiliary input port. This system uses an RGB-D camera and an Arduino interface board to capture motion data, including visual odometry and joystick signals, via ROS communication. Future motion is predicted using an autoregressive sparse Gaussian process model. We evaluate the proposed system on real-world short-term path prediction experiments. Experimental results demonstrate the system's efficacy when compared to a baseline neural network model.

\end{abstract}

\setlength{\parskip}{0.4em} 
\section{Introduction}\label{sec:intro}

\setlength{\parskip}{0.5em} 
It is estimated that by the year 2050 the number of people over the age of 85 will have tripled~\cite{desa2013world}, and a significant portion of the aging population is expected to need mobility
assistance. Confidence in independent mobility is core to psychological functioning \cite{anderson2013role}, and a greater sense of control can be positively correlated with a reduced mortality rate; consequently, the ability of powered wheelchairs (PWCs) to provide improved mobility could lead to a host of positive outcomes for this growing but mobility challenged population.  Unfortunately, older adults often have sensory, motor and/or cognitive impairments that preclude them from safely operating these large, heavy and powerful machines \cite{fehr2000adequacy}.

Smart wheelchairs (SWCs) seek to overcome the limitation of PWCs by using sensing, planning and control techniques from the robotics community to ensure safe operation and support the operator to accomplish tasks of daily living.  Constructing the ``smarts'' as an add-on to a commercially available PWC platform avoids a host of seating, mechanical design and reliability issues, but requires methods of predicting and controlling the motion of a device that was never intended for robotics.  Although commercial PWCs use power electronics, control systems, communication buses and input devices from several companies, the leading manufacturers consider their technical specifications and interfaces proprietary.  Furthermore, modifications to mobility critical aspects of the PWC, such as the drive train, are undesirable because of their complexity and the risk of a failure stranding the driver; consequently, mounting sensors on the drive wheels' axle to get wheel odometry is often impractical.

These complications mean that the basic capabilities assumed by most algorithms for wheeled robots---estimating what path the robot will take based on what motion commands were given, and estimating what motion commands will lead the robot along a particular path---are difficult to achieve on commercial PWC platforms.  It is, of course, still possible (and desirable) to do low latency reactive collision avoidance without these prediction capabilities, but users and bystanders prefer to avoid the abrupt jerks associated with such reactions when possible.  For the SWC, we want predictive capabilities in order to (a) reduce collision likelihood with earlier (and hopefully more subtle) interventions and (b) synthesize interventions designed to help the user achieve desired destinations over longer time periods.

In order to capitalize on the autonomous motion techniques developed by the robotics community and thereby turn a PWC into an SWC, we propose to provide these basic prediction capabilities using a combination of visual odometry, a standard analog ``alternative input'' port on the PWC, and autoregressive sparse Gaussian process models.  The visual odometry is obtained through a low-cost, lightweight and discretely mounted RGB-D camera.  An Arduino board with a simple custom shield card provides the interface between the PWC and a laptop running ROS.  We evaluate the system in real-world environments, including different floor and lighting conditions. Experimental results demonstrate the efficacy of our estimates.  Because this infrastructure can be easily ported to a wide variety of PWC bases from different manufacturers and integrated with a wide variety of sensing and planning technology available in ROS, we believe it represents a significant step toward bringing SWC autonomy into the PWC marketplace.

\section{Related Work}\label{sec:related_work}

\subsection{Smart Powered Wheelchairs}
Research on SWCs has a long history and is still an active area. An early survey can be found in \cite{simpson2005smart}, so here we constrain the discussion to some recent work. In \cite{doshi2007efficient} the authors take a Bayesian approach to learning the user model simultaneously with a dialog manager policy for intelligent planning. PerMMA \cite{cooper2012personal} combines manipulation and mobility assistance in support of complete independence for its users. A wizard of oz experiment (in which a human teleoperator simulates the SWC) was used in \cite{viswanathan2017intelligent} to explore SWC control strategies for older adults with cognitive impairment, including user attitudes, needs, and preferences. Seating pressure sensors are used to monitor the user in \cite{ma2016activity}.  An assessment of driving assistance by a deictic command for an SWC is proposed in \cite{leishman2014driving}, which enables the user to indicate desired motion on an interface displaying a view of the environment. An SWC capable of autonomous navigation in urban environments which integrates 3D perception is developed in \cite{schwesinger2017smart}.  However, all of this work has focused on other features of an SWC, such as autonomous navigation, collision avoidance or detection of abnormal user behavior; little work has been focused on motion prediction for general PWC platforms.

\subsection{Visual Odometry}\label{sec:fovis}
Visual odometry (VO) is the process of estimating the ego-motion of an agent using as a measurement input the images from cameras attached to it. Compared with wheel odometry, VO is not affected by the wheel slip common in uneven terrain or slippery environments. It has been demonstrated that VO can generate more accurate trajectory estimates than wheel odometry, with relative position error ranging from $0.1\%$ to $2\%$ \cite{scaramuzza2011visual}.
Sparse feature-based methods \cite{nister2004visual,huang2011visual} and dense photometric-error based methods \cite{kerl2013robust} are two widely used methods for VO. The former is based on salient and repeatable features that are tracked over the frames, while the latter uses the intensity information of all the pixels in the image. In this work, we chose to use the sparse feature-based VO package Fovis \cite{huang2011visual} based on a comparison~\cite{fang2014experimental} with five other VO algorithms (including DVO \cite{kerl2013robust} and GICP \cite{segal2009generalized}): Fovis was shown to have just half the runtime and half the average CPU usage; additionally, it performed best in complex environments with long corridors and cluttered or spacious rooms.

\subsection{Learning Robot Motion Models}
Machine learning methods have been widely applied to robotics tasks such as manipulation \cite{levine2016end}, autonomous driving \cite{bojarski2016end} and localization \cite{Lili_IROS2017}, and we propose to similarly learn the motion model of the PWC.  This exact problem was studied in \cite{talebifard2014risk}, which built a simple feed-forward neural network with a single hidden layer and only first-order differential information from the joystick and odometry readings.  A more complex probabilistic odometry model from input commands is presented in \cite{eliazar2004learning}, but their motion model is mainly developed for use in a simultaneous localization and mapping (SLAM) algorithm. An FCN-LSTM architecture is used in \cite{xu2016end} to learn a driving model from a large scale outsourced video dataset with GPS labels; however, it uses manual labeling and GPS signals only available in outdoor driving scenarios, while our focus remains on the indoor environments more frequently encountered by older adults.

\section{The PWC Hardware and Software Platform}\label{sec:platform}

\begin{figure}[h]
\centering
\includegraphics[width=0.7\linewidth]{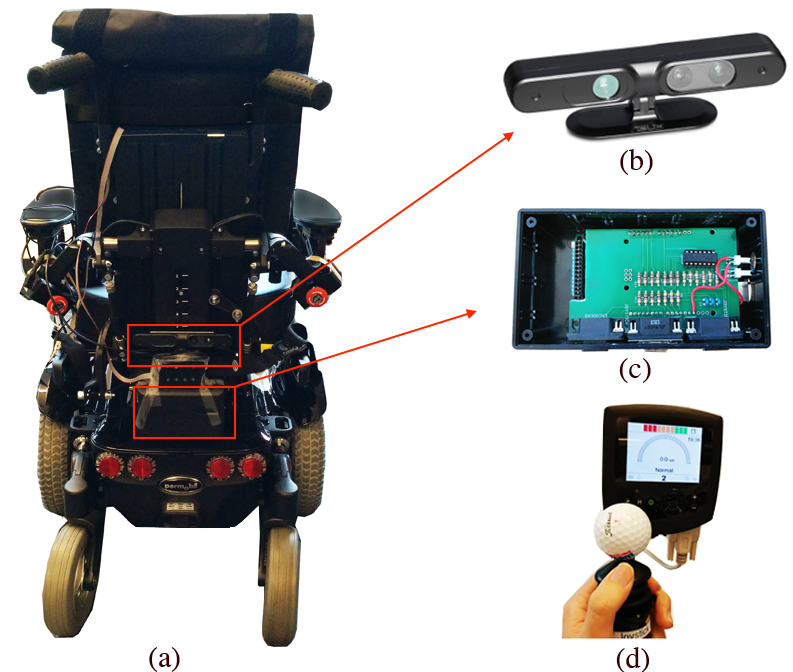}
\caption{\textbf{PWC hardware platform}. (a) The back view of our PWC.  (b)  An ASUS Xtion RGB-D camera used to capture the visual odometry data. For convenience, we installed it at the back during experiments; it could be installed elsewhere. (c) Arduino control box (opened to show contents). (d) A JC200 analog joystick used to control the PWC through the R-net Omni+ alternative input port. }
\label{fig:wheelchair}
\vspace{-3mm}
\end{figure}

The main hardware components of our platform are shown in Fig. \ref{fig:wheelchair}.  The PWC is an off-the-shelf commercial Permobil M300 Corpus 3G.  The PWC's built-in joystick is overridden by a Rnet Omni+ alternative input device port into which we can plug a Penny \& Giles JC 200 joystick.  To intercept and/or modify the joystick signal, we use a custom analog interface board attached as a shield to an Arduino Mega single-board microcontroller.  The custom board can read the analog joystick and write signals that look like the analog joystick's signals to the Omni+ input device port.  The board is controlled by the Arduino, which in turn communicates through a serial port with a Lenovo W530 laptop (Intel Core i7 with 8GB memory) running ROS.  An Asus Xtion RGB-D camera is mounted looking backward behind the seat to avoid interference with the driver and to reduce the aesthetic impact; the camera could be installed elsewhere. The camera can deliver RGB frames at 30Hz and depth images with $640 \times 480$ resolution and $58^\circ$ HFV.

 \begin{figure}[h]
 \centering
 \begin{minipage}{1\linewidth}
 \centering
 \includegraphics[width=\textwidth]{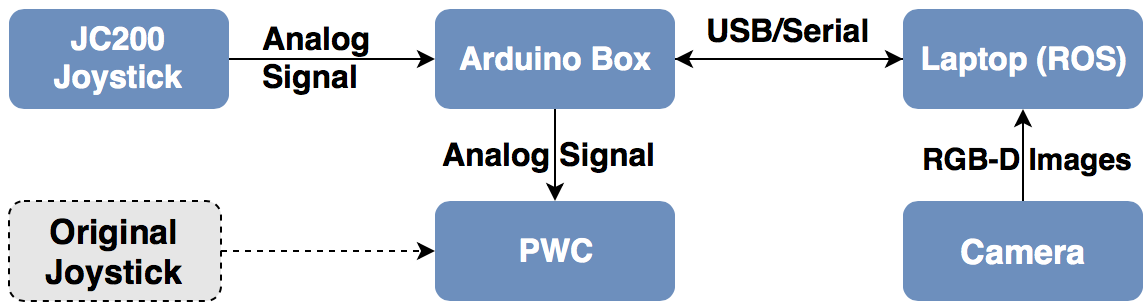}
 {(a) }
 \end{minipage} 
 \begin{minipage}{1.0\linewidth}
 \centering
\includegraphics[width=\textwidth]{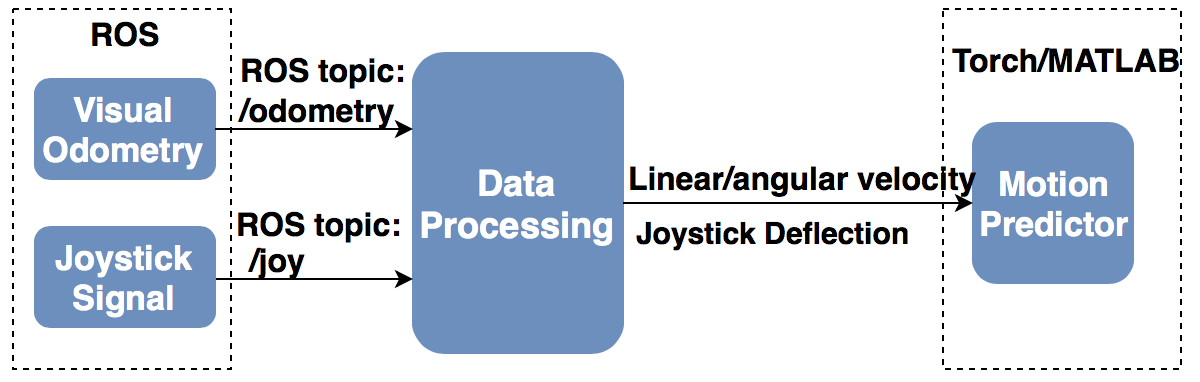}
 {(b) }
 \end{minipage} 
 \vspace{+0mm}
\caption{\textbf{Our communication pipelines.} (a) Hardware pipeline, (b) Software pipeline.}
\label{fig:system}
\end{figure} 

Fig. \ref{fig:system} (a) shows how the hardware components interact:
\begin{compactitem}
  \item The joystick sends a control signal to the Arduino.
  \item The Arduino converts the control signal to digital form and passes it to the laptop running ROS for prediction and/or logging.  Although not used here, the laptop can also send a control signal back to the Arduino.
  \item The Arduino converts its digital signal back to analog and sends it to the PWC to override the built-in joystick (shown in the dashed box).
  \item The RGB-D camera passes images to the laptop.
\end{compactitem}

The software pipeline is shown in Fig. \ref{fig:system} (b). During data collection, the joystick and VO data on the indicated ROS topics are stored in a bag file on the laptop.  After collection, the data are resampled using linear interpolation onto a common 5Hz frequency, and the resampled data are used to train or test the motion prediction model.

\section{Autoregressive Sparse Gaussian Processes}\label{sec:methods}
In this section, we introduce our autoregressive sparse Gaussian process model for motion prediction. While we seek a pose trajectory, VO provides velocity data so we will estimate the velocity trajectory and integrate to get pose.

We denote an element of our time series by the joystick and velocity pair $(J_t, V_t)\in \mathbb{R}^4$,  where $J_t \in \mathbb{R}^2$ and $V_t \in \mathbb{R}^2$ represent the joystick deflections and velocity at time step $t$ respectively. The components of $J_t$ are the forward-backward deflection (intuitively the linear velocity command) and leftward-rightward deflection (intuitively the angular velocity command).  The components of $V_t$ are linear and angular velocity measurements which we will denote as $[v_t, \omega_t]^T$.   Inspired by the time series autoregressive process \cite{brockwell2013time}, we use the past $s$ time-steps of the time series as the input to predict the current time-step $(J_t, V_t)$. In fact, because we are concerned only with velocity prediction, we also provide the current joystick $J_t$ to our prediction function.  Defining
\[ \label{eq:autostate}
\begin{aligned} 
X_t &= \{\bar{X}_t, J_t \},\\
\bar{X}_t &=[J_{t-s},\dots,J_{t-1},V_{t-s},\dots,V_{t-1}]^T, \nonumber
\end{aligned}
\]
we can for notational convenience write our prediction function in several different ways:

\begin{equation}  \label{eq:mainauto}
\begin{aligned}
\hat{V}_{t} &= f(J_{t-s}, \dots, J_{t-1}, V_{t-s}, \dots, V_{t-1}, J_t),\\
  &= f(\bar X_t, J_t), \\
  &= f(X_t), \\
V_t &= \hat{V}_t + \epsilon_t,
\end{aligned}
\end{equation}
where $\epsilon_t$ represents a noise term that can not be learned. 

\begin{figure}
\centering
\includegraphics[width=\linewidth]{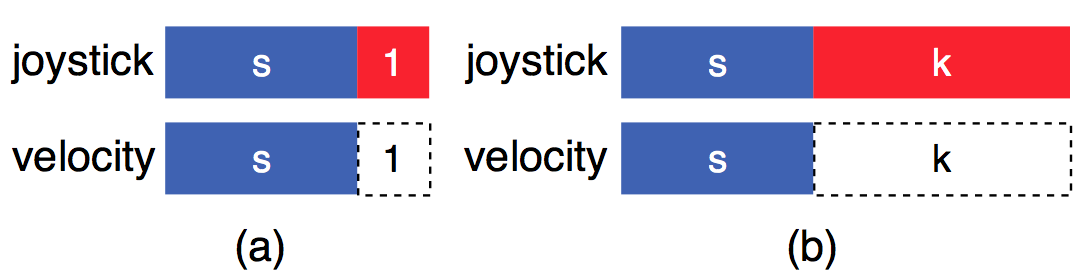}
\vspace{-4mm}
\caption
{\textbf{Time-series velocity prediction.} (a) Immediate velocity prediction. At time-step $t$, given the previous $s$ time-steps of both joystick and velocity, and the current time-step joystick input $J_t$, predict the velocity $V_t$ (the dashed part).  (b) Velocity sequence prediction. Given the past $s$ time-steps of both joystick and velocity, and the future $k$ time-steps joystick input 
     ($J_t, J_{t+1}, \cdots, J_{t+(k-1)}$), predict the future k time-steps velocity ($V_t, V_{t+1}, \cdots, V_{t+(k-1)}$). }
\label{fig:prediction}
\end{figure}

\begin{algorithm}
\caption{Predicting velocity sequence. }
\label{alg:k_predict}
\begin{algorithmic}[1]
\Require a past velocity sequence $\{V_{t-s},\cdots ,V_{t-1} \}$.
\Require a past and future joystick command sequence $\{J_{t-s},\cdots ,J_{t-1},J_{t}, J_{t+1},\cdots ,J_{t+(k-1)}  \}$.
\Ensure A sequence of velocities $\{\hat{V}_{t}, \hat{V}_{t+1} \cdots, \hat{V}_{t+(k-1)}\}$  
\State seq = \text{[ ]}
\For{\text{i= 0 to k-1}}
\State $\hat{V}_{t+i} = f(\bar{X}_{t+i}, J_{t+i})$
\State Update the unknown $V_{t+i}$ with $\hat{V}_{t+i}$ in $\bar{X}_{t+i+1}$
\State seq.push($\hat{V}_{t+i}$)
\EndFor
\State \Return  seq
\end{algorithmic}
\end{algorithm}

We employ autoregressive sparse Gaussian processes (ASGP) as the motion prediction model.  An ASGP integrates an autoregressive process with a sparse Gaussian process (SPGP) using pseudo-inputs \cite{snelson2006sparse}. SPGP is a Gaussian process regression model whose training cost is $O(M^2N)$ and prediction cost is $O(M^2)$, where $M$ is the number of pseudo-inputs, $N$ is the number of training samples, and $M\ll N$.

Algorithm \ref{alg:k_predict} provides the details of the prediction process. Given an additional $k$ time-steps of future joystick commands, the Recurrent Sliding Window method \cite{dietterich2002machine} uses the predicted velocity as the input for predicting the next velocity until a $k$-step velocity sequence is predicted (as shown in Fig. \ref{fig:prediction}). The interface for velocity sequence prediction is summarized by :
\begin{equation} \label{eq:traject_vel}
\begin{aligned}
(\hat{V}_{t},\ldots,&\hat{V}_{t+(k-1)}) \\
    & = f_V(J_{t-s},\dots, J_{t+(k-1)}, V_{t-s}, \dots, V_{t-1})
\end{aligned}
\end{equation}
where $\hat{V}_{t},\ldots,\hat{V}_{t+(k-1)}$ are the predicted velocity sequence of length k. In practice, the future joystick inputs $J_{t+1}, \ldots, J_{t+k-1}$ can be obtained from a path planner. 

We will note in passing that since $V_t = [v_t, \omega_t]^T \in \mathbb{R}^2$, Eq. \ref{eq:mainauto} actually represents two separate prediction functions:
\begin{align}
\hat{v}_{t} &= f_v(j_{t-s},\dots, j_{t-1}, v_{t-s}, \dots, v_{t-1}, j_{t}) \label{eq:linear_imm} \\ 
\hat{\omega}_{t} &= f_\omega(j_{t-s},\dots, j_{t-1},\omega_{t-s}, \dots, \omega_{t-1}, j_{t})\label{eq:angular_imm}
\end{align}
where $j_i$ in the two equations are the corresponding linear (for $f_v$) or angular (for $f_\omega$) joystick input.

Given the velocity sequence $\{ V_i \}_{i=1}^k$, we integrate using a unicycle model
\begin{equation} \label{eq:unicycle_de}
  \begin{bmatrix}
    \dot x \\ \dot y \\ \dot \theta 
  \end{bmatrix}
    =
  \begin{bmatrix}
    v \cos \theta \\ v \sin \theta \\ \omega
  \end{bmatrix}
\end{equation}
to obtain a corresponding pose trajectory $\{ P_i \}_{i = 1}^k = \{ (x_i, y_i, \theta_i) \}_{i=1}^k$ where $(x_i,y_i)$ and $\theta_i$ are the position and the heading of the wheelchair respectively. Since discrete time data is provided, we use a trapezoidal quadrature to approximate the integral.

\section{Experiments}\label{sec:experiment}
In this section, we evaluate our system in real-world experiments with different lighting and floor conditions. 

\subsection{Data Collection}\label{sec:dataset}

\begin{figure}
\centering
\begin{minipage}{1\linewidth}
\centering
\includegraphics[width=\linewidth]{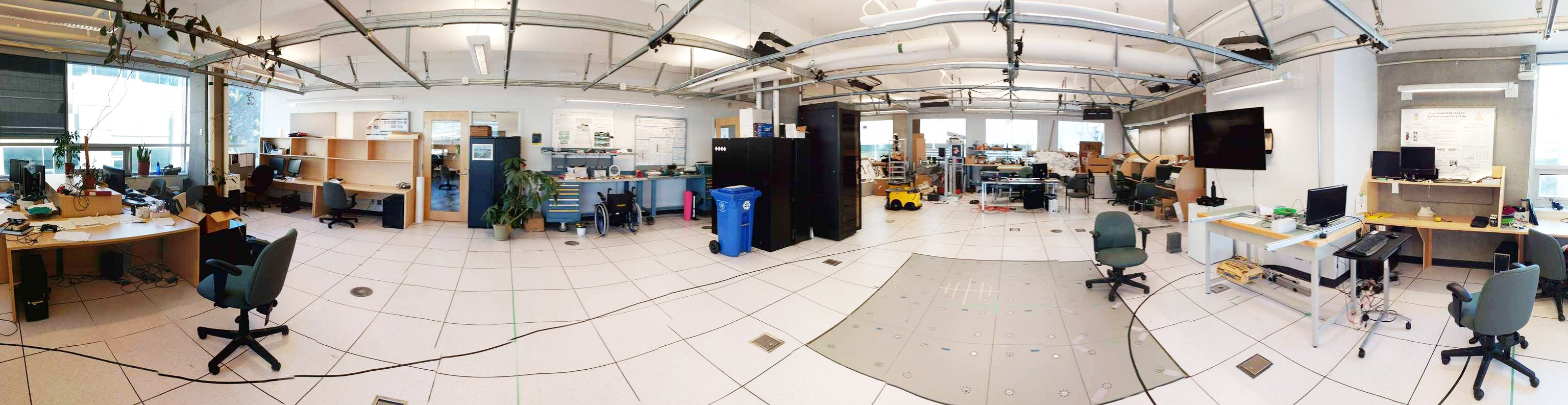}\\
(a)
\end{minipage}
\begin{minipage}{1\linewidth}
\centering
\includegraphics[width=\linewidth]{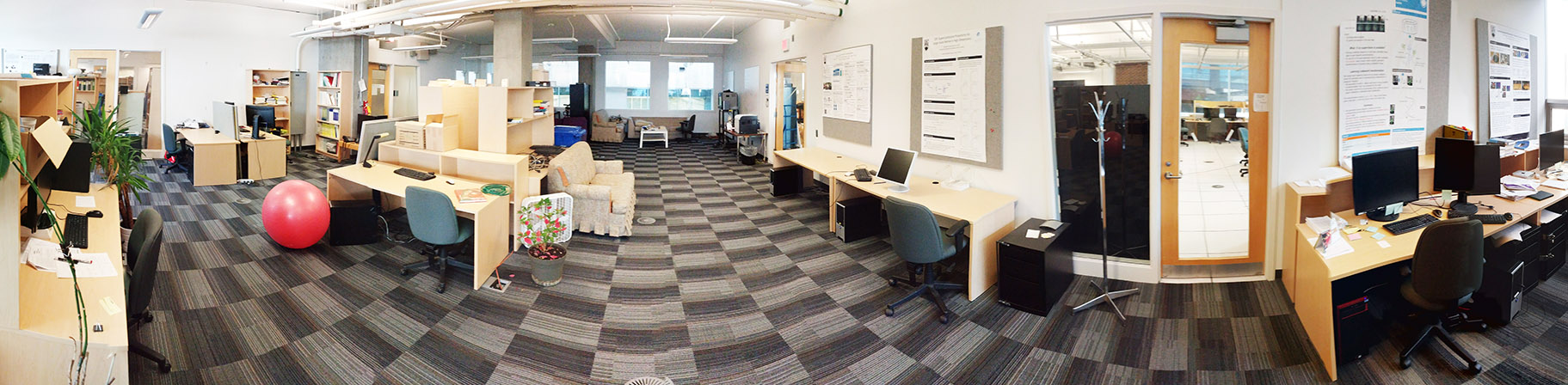} \\
(b)  
\end{minipage}  
\begin{minipage}{1\linewidth}
\centering
\includegraphics[width=\linewidth]{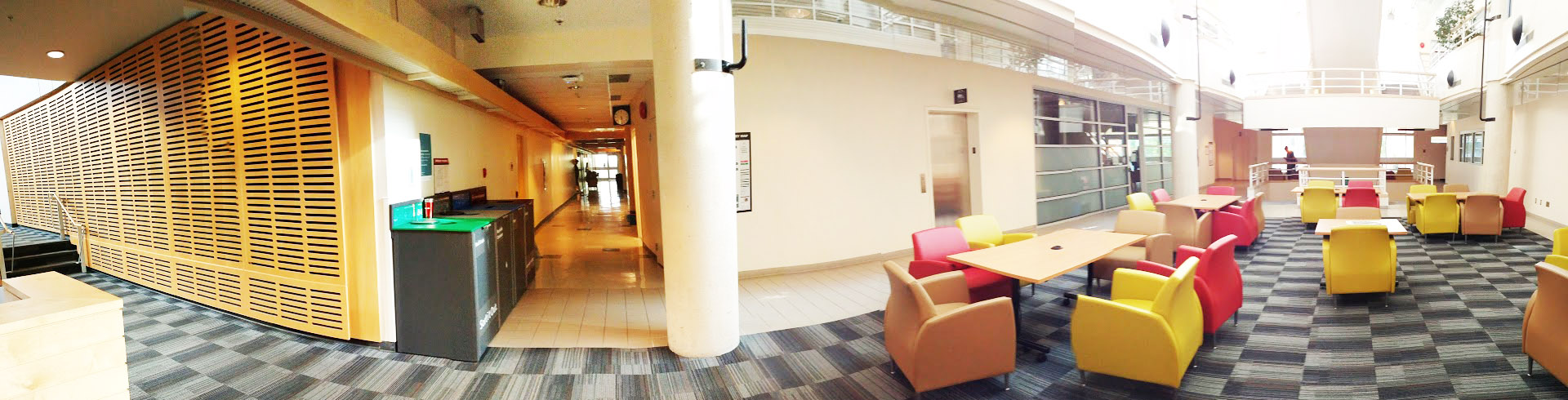} \\
\vspace{-1mm}
(c)  
\end{minipage}  
\caption{\textbf{Panorama view of experimental environment.} Best viewed in color. (a) lab with tiles, (b) lab with carpets, (c) public area with both carpet and tile flooring (``hybrid''). }
\label{fig:lab}
\end{figure}

In the data collection stage, we drive the wheelchair around our experimental environments attempting to manually generate roughly random (but safe) trajectories.  The three indoor environments shown in Fig. \ref{fig:lab} were used:
\begin{compactenum}
\item A lab with only tile floor.
\item A lab with only carpet floor.
\item An atrium area with both (the same) carpet and (different) tiles.  We call this the ``hybrid'' environment.
\end{compactenum}

We collected data in eight ROS bags for environments 1 and 2 respectively. To evaluate our model's performance on a complex and not previously encountered environment, two bags were collected from environment 3.  Each bag contains about 20 minutes of sequential data, and the data sequences in each begin and end with the wheelchair motionless and the joystick zeroed for several seconds so that they can be concatenated in arbitrary order.  In all of the analysis below, we use six bags (120 min) of data for training and two bags (40 min) for testing.

In the discrete time paradigm of ASGP, all inputs must be delivered simultaneously to the regression models. Unfortunately, the VO data arrives at a very jittery 4--7Hz whereas the joystick data arrives more consistently at roughly 30Hz.  In order to provide the required simultaneous input data for the regression model we linearly interpolate the raw joystick and velocity data points and resample at 5Hz.

One added complication is that the VO provided by Fovis occasionally fails in situations where very few matching feature points are detected. During these periods Fovis continues to report the last measured velocity, which can lead to incorrect motion estimates.  In practice, we did not see any such failures lasting more than a few seconds.

\subsection{Model Details}
To compare against the ASGP model of Eq. \ref{eq:mainauto} we use a Multilayer Perceptron (MLP) model similar to \cite{talebifard2014risk} implemented in the Deep Learning framework Torch \cite{collobert2011torch7}.  We use three hidden linear layers each containing 30 neurons, and ReLU layers are used to connect the three linear layers. The input and output are the same as ASGP.  We chose a batch size of 100 to train using the Adam optimization method \cite{kingma2014adam}. 

The hyperparameter $s$ specifying the number of past steps available to the model in Eq. \ref{eq:mainauto} was manually tuned to be $s=4$ for ASGP and $s=10$ for MLP based on results from the tile training set.  Finally, we manually tuned the number of pseudo-inputs used by SPGP as 20 for predicting the linear velocity and 40 for predicting the angular velocity based on results from the tile training set.  Although additional pseudo-inputs continued to improve results, the benefits were marginal beyond these values and the cost (both training and evaluation) grows quadratically.  The parameter $k$ in Eq. \ref{eq:mainauto} was determined by the desired prediction horizon and the sampling frequency of 5Hz: For a horizon of 1000ms $k = 5$, for 1400ms $k = 7$ and for 2000ms $k = 10$.

\subsection{Results}\label{sec:result}
We train separate ASGP and MLP models to predict linear and angular velocity trajectories and then we integrate these velocities to generate pose trajectories out to horizon $k$.  These predicted pose trajectories are compared with pose trajectories derived by integration of the recorded VO for the same horizon.  A variety of error measurements are possible, but for reasons of space, we focus on the L2 norm of the difference between the predicted $(x, y)$ position and the recorded position.

To provide a quantitative comparison between the models for a variety of training and test data set combinations, we examine the percentage of pose predictions from the test set with the error no larger than 10cm (corresponding roughly with the mapping error encountered when using common SLAM algorithms and RGB-D cameras for indoor occupancy grid construction).  We will call this percentage the ``success rate'' of the model for a given training and test dataset pair.  Results are shown in Tables \ref{table:pose_traject_1000ms}, \ref{table:pose_traject_1400ms} and \ref{table:pose_traject_2000ms} for 1.0, 1.4 and 2.0 second horizons respectively.  For each horizon, we train six different models: an MLP and an ASGP model using each of
\begin{compactitem}
\item ``Tile'': 120 minutes of tile data Fig. \ref{fig:lab}(a).
\item ``Carpet'': 120 minutes of carpet data Fig. \ref{fig:lab}(b). 
\item ``Both (1:1)'': A combined training set with 60 minutes of tile and 60 minutes of carpet data.
\end{compactitem}
We then evaluate each of the six models against three different test sets
\begin{compactitem}
\item ``Tile'': 40 minutes of tile data Fig. \ref{fig:lab}(a). 
\item ``Carpet'': 40 minutes of carpet data Fig. \ref{fig:lab}(b). 
\item ``Hybrid'': 40 minutes of data from the novel atrium environment Fig. \ref{fig:lab}(c). 
\end{compactitem}
For example, the top left value in each table corresponds to the percentage of tile test samples whose prediction error was smaller than 10cm for the MLP model trained on the tile data set. Turning our attention specifically to Table \ref{table:pose_traject_1000ms}, the top left cell shows that the MLP model trained on a 120 minute tile data set (Fig. \ref{fig:lab} a) and tested on a 40 minute tile data set showed a 92.21\% success rate, corresponding to approximately 11,065 test cases with L2 position estimate error (at a 1 second horizon) smaller than 10cm out of (40 minutes $\times$ 60 seconds / minute $\times$ 5 samples / second) = 12,000 test cases. 

\begin{table}
\caption{{\upshape Percent of pose prediction with no more than 10 cm error at 1000ms horizon.}}
\label{table:pose_traject_1000ms}
\begin{tabular}{|c|c|c|c|c|}
\hline
\multicolumn{2}{|c|}{\backslashbox{\textbf{Test} (40 min)}{\textbf{Train} (120 min)}} & \textbf{Tile} & \textbf{Carpet} & \textbf{Both} (1:1) \\ \hline
\multirow{2}{*}{\textbf{Tile}} & MLP & 92.21 & 92.66 & 91.69 \\ \cline{2-5} 
 & ASGP & 94.86 & 94.55 & 94.51 \\ \hline
\multirow{2}{*}{\textbf{Carpet}} & MLP & 82.21 & 75.26 & 83.45 \\ \cline{2-5} 
 & ASGP & 83.79 & 82.12 & 82.94 \\ \hline
\multirow{2}{*}{\textbf{Hybrid}} & MLP & 79.96 & 82.04 & 83.46 \\ \cline{2-5} 
 & ASGP & 84.01 & 83.05 & 82.35 \\ \hline
\end{tabular}
\end{table}

\begin{table}
\centering
\caption{{\upshape Percent of pose prediction with no more than 10 cm error at 1400ms horizon.}}
\label{table:pose_traject_1400ms}
\begin{tabular}{|c|c|c|c|c|}
\hline
\multicolumn{2}{|c|}{\backslashbox{\textbf{Test} (40 min)}{\textbf{\textbf{Train}} (120 min)}} & \textbf{Tile} & 
\textbf{Carpet} & \textbf{Both} (1:1) \\ \hline
\multirow{2}{*}{\textbf{Tile}} & MLP & 81.07 & 79.37 & 75.03 \\ \cline{2-5} 
 & ASGP & 85.91 & 85.01 & 85.03 \\ \hline
\multirow{2}{*}{\textbf{Carpet}} & MLP & 67.66 & 50.56 & 69.57 \\ \cline{2-5} 
 & ASGP & 69.40 & 68.72 & 69.66 \\ \hline
\multirow{2}{*}{\textbf{Hybrid}} & MLP & 63.61 & 63.55 & 67.35 \\ \cline{2-5} 
 & ASGP & 68.50 & 68.73 & 67.72 \\ \hline
\end{tabular}
\end{table}

\begin{table}
\centering
\caption{{\upshape Percent of pose prediction with no more than 10 cm error at 2000ms horizon.}}
\label{table:pose_traject_2000ms}
\begin{tabular}{|c|c|c|c|c|}
\hline
\multicolumn{2}{|c|}{\backslashbox{\textbf{Test} (40 min)}{\textbf{Train} (120 min)}} & \textbf{Tile} & \textbf{Carpet} & \textbf{Both} (1:1) \\ \hline
\multirow{2}{*}{\textbf{Tile}} & MLP & 63.14 & 60.60 & 52.06 \\ \cline{2-5} 
 & ASGP & 72.37 & 70.64 & 71.19 \\ \hline
\multirow{2}{*}{\textbf{Carpet}} & MLP & 43.79 & 31.11 & 51.00 \\ \cline{2-5} 
 & ASGP & 50.21 & 51.69 & 51.69 \\ \hline
\multirow{2}{*}{\textbf{Hybrid}} & MLP & 40.76 & 41.82 & 43.48 \\ \cline{2-5} 
 & ASGP & 47.91 & 50.89 & 48.77 \\ \hline
\end{tabular}
\end{table}

\subsection{Discussion}

Tables \ref{table:pose_traject_1000ms}, \ref{table:pose_traject_1400ms} and \ref{table:pose_traject_2000ms} show that ASGP convincingly outperforms MLP for predicting future position.
Averaging across all combinations of training and test data sets, ASGP outperforms MLP by $2.14\%$, $5.66\%$, $9.73\%$ for 1000 ms, 1400 ms and 2000 ms horizon predictions respectively, and in only two cases does MLP beat ASGP: Trained on ``Both'' and tested on ``Carpet'', the success rate at a 1000 ms horizon of MLP is $83.45\%$ and of ASGP is $82.94\%$; trained on ``Both'' and tested on ``Hybrid'', the success rate of MLP is $83.46\%$ and of ASGP is $82.35\%$. However, ASGP regains an advantage at longer horizons.

Interestingly, ASGP also seems to be more robust to variation in training data: Different training data sets do not affect the success rate for different test sets nearly as dramatically as with MLP.  For example, in Table \ref{table:pose_traject_1000ms}, when tested on the tile, carpet or hybrid data sets, ASGP has a success rate of about $94\%$, $83\%$ and $83\%$ respectively, regardless of which training data set was used.  The same pattern occurs in Tables \ref{table:pose_traject_1400ms} and \ref{table:pose_traject_2000ms}. Furthermore, the success rate of ASGP for testing on carpet and hybrid do not differ significantly at any fixed horizon, even though the latter is a novel environment: The success rate for testing on either carpet or hybrid at 1000 ms, 1400 ms or 2000 ms is about $83\%$, $68\%$ and $50\%$ respectively.  In contrast, the success rates of various MLP models vary widely for a given test data set depending on which training data set was used.

\begin{figure}
 \centering
 \begin{minipage}{0.20\textwidth}
 \centering
 \includegraphics[width=\textwidth]{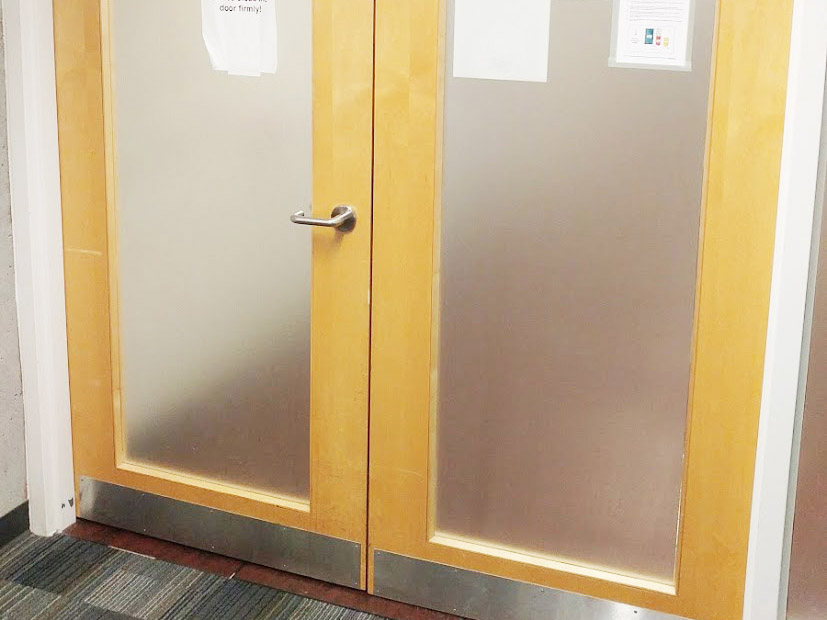}  
 {(a) }
 \end{minipage} 
  \begin{minipage}{0.20\textwidth}
 \centering
\includegraphics[width=\textwidth]{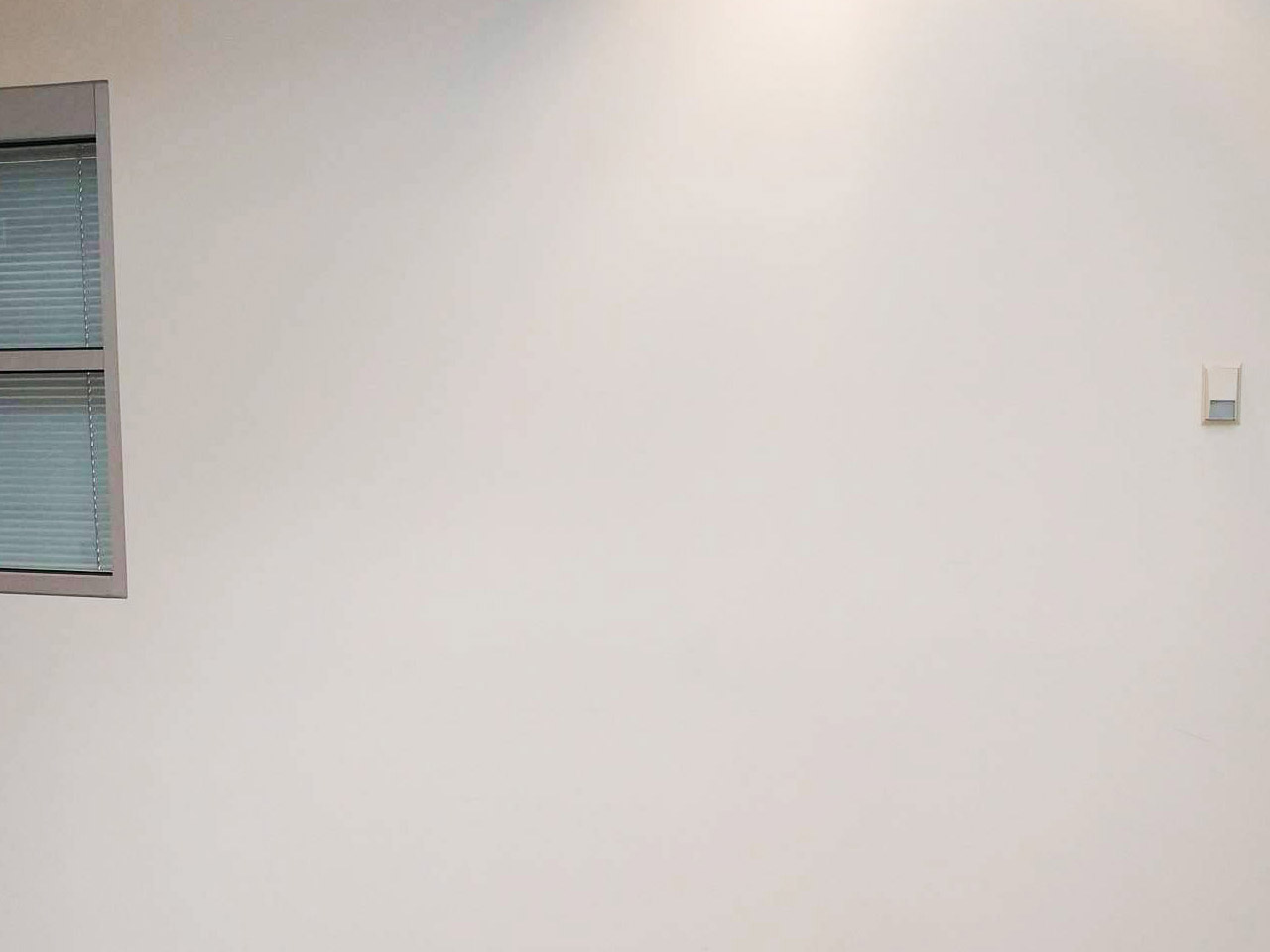} 
 {(b) }
 \end{minipage} 
 \vspace{-0mm}
\caption{\textbf{Locations in the data with limited visual features}: (a) carpet environment; (b) hybrid environment.}
\label{fig:no_vo_space} 
\end{figure} 

A final observation arising from Tables \ref{table:pose_traject_1000ms}, \ref{table:pose_traject_1400ms} and \ref{table:pose_traject_2000ms} is that success rates for testing on the tile data set are consistently higher than success rates for the other two test sets.  We hypothesize this result is due to better VO performance in the tile environment brought on by a combination of factors.  First, the tile environment has bright artificial lighting, whereas the carpet environment has dimmer artificial lighting and the atrium has patches of strong sunlight passing through a glass wall (as shown on the right side of Fig. \ref{fig:lab} (c)), both lighting situations which can lead to limited visual features \cite{munaro2013software}.  Second, the tile environment is cluttered but stable, while both the carpet and atrium environments have regions where few visual features are detected (see Fig. \ref{fig:no_vo_space} for examples).  These results again highlight the fact that VO is most effective in a rich visual environment.

\section{Conclusions}\label{sec:conclusion}
A key step in the creation of a smart wheelchair built on a commercial powered wheelchair (PWC) platform is the ability to predict the motion of the PWC.  In this paper, we presented an integrated hardware and software system to do so which requires no modifications to the PWC itself: We merely mount an RGB-D camera in an inconspicuous location and plug into a standard alternative input port.  We used an autoregressive sparse Gaussian process model to make pose trajectory predictions based on visual odometry and joystick data. The proposed system was evaluated in academic building environments with different floor surfaces and different richness of visual features. The results demonstrate the adequacy of our system in the test environment and superior performance to a neural network model.

In the future, we plan to enable our system to work online, to provide pose trajectory predictions with uncertainty estimates, and to implement the inverse of the current function to provide a mapping from trajectories to joystick sequences.  Our goal is to better predict where the user is steering the PWC, the likelihood of that steering choice progressing toward the user's desired destination or leading to a collision, and the most appropriate input intervention signal should that steering choice appear undesirable.

\section*{Acknowledgements}

The authors would like to thank Martin Gerdzhev and Professor Joelle Pineau's team at McGill for sharing their Arduino code and shield card to interface with the Omni+ input port, Emma Smith and Professor William C. Miller's team at UBC for their work securing the PWC platform, and Justin Reiher and previous members of the AGEWELL WP3.2 and CanWheel engineering teams for their work on the hardware and software systems used in this research.

  \addtolength{\textheight}{-11cm}  
  \bibliographystyle{abbrv}
  \bibliography{sigproc}  
\end{document}